# Letter: Can CNN Construct Highly Accurate Models Efficiently for High-Dimensional Problems in Complex Product Designs?

**Yu Li**[a, *], **Hu Wang**[a, **], **Juanjuan Liu** [a]

*a. State Key Laboratory of Advanced Design and Manufacturing for Vehicle Body, Hunan University, Changsha, 410082, P.R. China*

## Abstract

The main purpose of this letter is to draw attention to an interesting and amazing way to solve the "curse of dimensionality" in metamodeling problems by using Convolutional Neural Network (CNN). With the increase of the nonlinearity and dimension, it is difficult for the present popular metamodeling techniques to construct metamodels efficiently. In this letter, the potential ability of the CNN to handle strongly nonlinear and highly dimensional (hundreds of dimensions) problems is discussed, evaluated and compared with some other popular methods. Moreover, the CNN-based metamodel is also applied to an IsoGeometric Analysis (IGA)-based optimization application, and the results further validate the high-dimensional ability of the CNN-based metamodel.

*Keywords*: CNN-based metamodel; High-dimensional; Strong-nonlinear; Complex product designs

## 1. Introduction

Currently, there are multiple kinds of metamodeling techniques. However, they suffer from a well-known bottleneck in high-dimensional designs [1-3]. Actually, some original design problems, such as vehicle body structure and aircraft designs, possess more than one hundred design variables. A popular way to overcome the high dimension, most of these problems are generally decomposed of several subproblems. However, some decomposed problems are actually not weak correlative.

[*] First author. *E-mail address*: liyu_hnu@hnu.edu.cn (Y. Li)

[**] Corresponding author. Tel.: +86 0731 88655012; fax: +86 0731 88822051.
*E-mail address:* wanghu@hnu.edu.cn (H. Wang)

In order to tackle the high dimension, Shan and Wang [4] integrated the Radial Basis Function (RBF) with High Dimensional Model Representation (HDMR) into RBF-HDMR, which was the first metamodel to solve high-dimensional problems. In addition, to overcome the high-dimensional shortcoming of Kriging (KG), Bouhlel [1] combined the KG with the Partial Least Squares (PLS) technique and proposed KPLS. At the same time, Chen [5] developed a novel gradient-enhanced KG method which utilized only a partial set of gradients for high dimensional problems. Compared with other techniques, Sudjianto [6] and Friedman [7] applied the Multivariate Adaptive Regression Spline (MARS) to modeling and optimization problems. Nevertheless, although these methods achieved some improvements for high-dimensional problems, they might be powerless in more than one hundred dimensions.

In this study, the CNN is suggested and attempted to over hundred-dimensional problems. It is well known that some Machine Learning (ML) methods have been utilized to some other disciplinaries, e.g. topology optimizations [8-11] and heat transfer [12-15]. As mentioned by Hornik [16], "multilayer feedforward networks with as few as one hidden layer are indeed capable of universal approximation in a very precise and satisfactory sense". Furthermore, the input to the CNN is usually a matrix of pixels of an image, which easily contains hundreds or even thousands of pixels. Therefore, it is considered to be a promising model for more than one hundred dimensional problems.

The main contributions and advantages of this study lie in three-folds: i. It might be the first time to study over hundred-dimensional and strong-nonlinear problems; ii. Compared with existing methods which are commonly employed for high-dimensional problems, the CNN outperforms in tests; iii. The CNN-based metamodel is applied to a complex geometric optimization successfully, which is commonly difficult for popular metamodeling techniques in product design, practically.

## 2. The CNN Theory

As shown in Fig. 1, in the CNN, each unit in the layer receives inputs from a set of units located in a small neighborhood in the previous layer. With local receptive fields, neurons can extract elementary visual features. These features are then combined by the subsequent layers in order to detect higher-order features, and can be applied by forcing a set of units to have identical weight vectors. Units in a layer are organized in planes within which all the units share the same set of weights. This operation is equivalent to a convolution, followed by an additive bias and active function, hence name convolutional network [17].

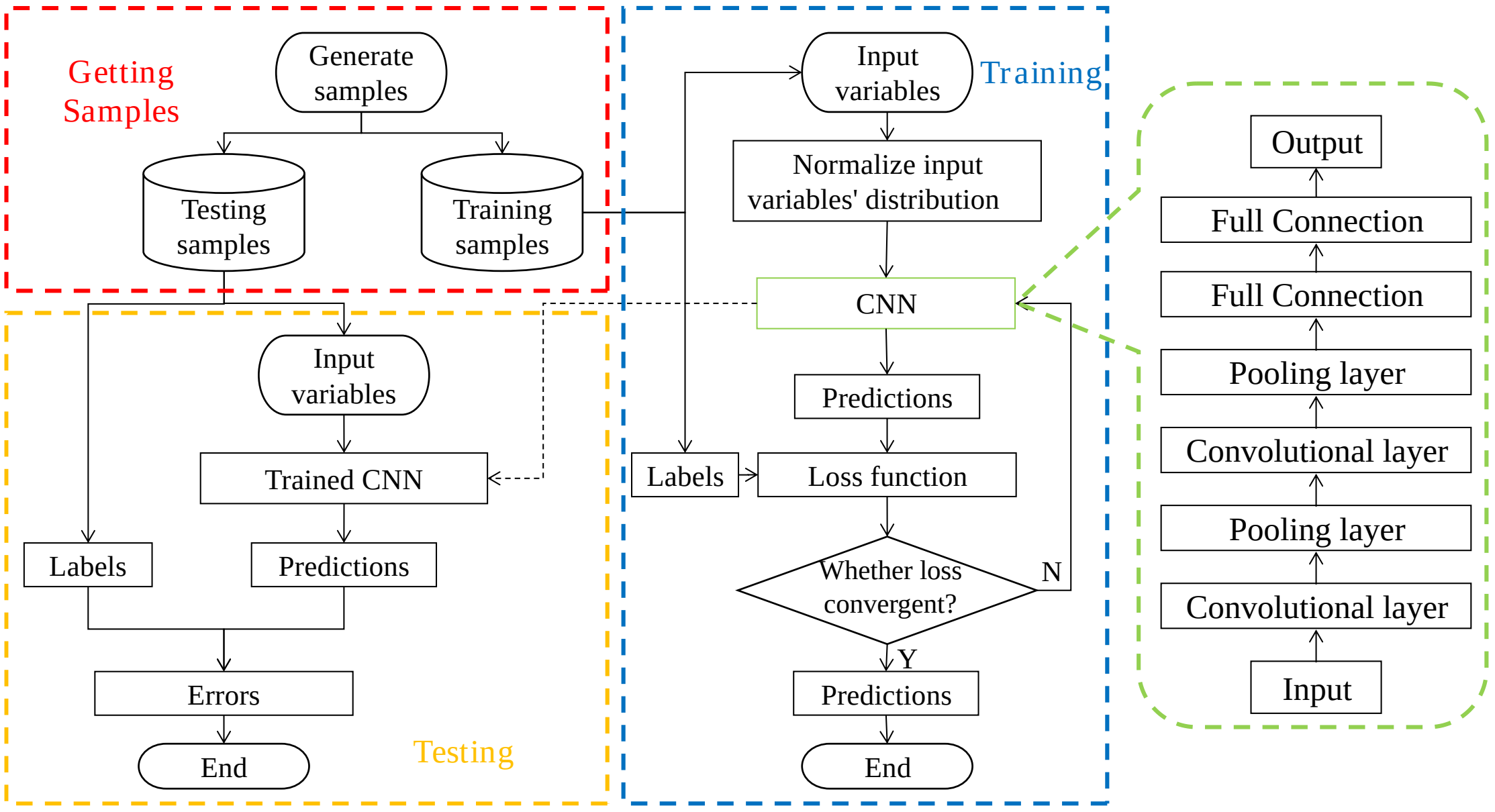

**Fig. 1.** The architecture of the CNN.

As for pooling layer, the forward propagation is done by a structure which is similar to the filter* in convolutional layers. However, the structure does not calculate the weighted sum, but a maximum or average value, which are well known as max pooling and average pooling.

The Mean Square Error (MSE) is usually used as a loss function to be minimized in regression problems. It can be represented as

* A $1\times1\times H$ or a $3\times3\times H$ matrix, and the $H$ is equal to the depth of input.

$$MSE(y,\hat{y}) = \frac{\sum_{i=1}^{n}(y_i - \hat{y}_i)^2}{n} \tag{1}$$

where $y_i$ is the actual label; $\hat{y}_i$ is the predicted value; and $n$ is the number of samples.

In this study, the optimization algorithm is Adaptive Moment Estimation (Adam) [18]. Mathematically, Adam can be defined as

$$g \leftarrow +\frac{1}{m}\nabla_\theta \sum_i L\left(f\left(x_i;\theta\right), y_i\right) \tag{2}$$

$$s \leftarrow \rho_1 s + (1-\rho_1) g \tag{3}$$

$$r \leftarrow \rho_2 r + (1-\rho_2) g \odot g \tag{4}$$

$$\hat{s} \leftarrow \frac{s}{1-\rho_1} \tag{5}$$

$$\hat{r} \leftarrow \frac{r}{1-\rho_2} \tag{6}$$

$$\Delta\theta = -\varepsilon \frac{\hat{s}}{\sqrt{\hat{r}} + \delta} \tag{7}$$

$$\theta \leftarrow \theta + \Delta\theta \tag{8}$$

where $L(x)$ is the loss function, $\theta$ is the initial parameter, $x_i$ is the training sample and $y_i$ is corresponding label, $m$ is the sample size, $s$ and $r$ are the first and second moment estimations respectively, $\rho_1$ and $\rho_2$ are the attenuation coefficients, and $\varepsilon$ is the learning rate. In this study, $\delta$=$10^{-8}$, $\rho_1$=0.9, and $\rho_2$=0.999.

## 3. Feasibility Analyses for Over One Hundred-Dimensional Problems by the CNN

### 3.1. Analytical examples

To evaluate the CNN-based metamodel performance, several typical high-dimensional and strong-nonlinear mathematical functions are tested.

**i. Griewank** [19] (interval: [-600, 600])

$$f_1(x)=\sum_{i=1}^{d}\frac{x_i^2}{4000}-\prod_{i=1}^{d}\cos\frac{x_i}{\sqrt{i}}+1 \tag{9}$$

**ii. Levy** [20] (interval: [-10, 10])

$$f_2(x)=\sum_{i=1}^{d}\left\{(x_i-1)^2\left[1+\sin^2(3\pi x_i)\right]\right\} \tag{10}$$

**iii. Weierstrass** [21] (interval: [-1, 1])

$$f_3(x)=\sum_{i=1}^{d}\left(\sum_{k=0}^{k\max}\left[a^k\cos\left(2\pi b^k(x_i+0.5)\right)\right]\right)-d\sum_{k=0}^{k\max}\left[a^k\cos\left(2\pi b^k\cdot 0.5\right)\right] \tag{11}$$

where $a$=0.5; $b$=3; and $k_{max}$=20.

**iv. Bent Cigar** [21] (interval: [-10, 10])

$$f_4(x)=x_1^2+10^6\sum_{i=2}^{d}x_i^2 \tag{12}$$

**v. Rotated Hyper-Ellipsoid** [19] (interval: [-5, 5])

$$f_5(x)=\sum_{i=1}^{d}\sum_{j=1}^{i}x_j^2 \tag{13}$$

**vi. Rosenbrock** [21] (interval: [-10, 10])

$$f_6(x)=\sum_{i=1}^{d-1}\left[100\left(x_i^2-x_{i+1}\right)^2+(x_i-1)^2\right] \tag{14}$$

**vii. HGBat** [21] (interval: [-100, 100])

$$f_7(x)=\left|\left(\sum_{i=1}^{d}x_i^2\right)^2-\left(\sum_{i=1}^{d}x_i\right)^2\right|^{1/2}+\left(0.5\sum_{i=1}^{d}x_i^2+\sum_{i=1}^{d}x_i\right)\Big/d+0.5 \tag{15}$$

**viii. Sum Squares** [19] (interval: [-10, 10])

$$f_8(x)=\sum_{i=1}^{d}ix_i^2 \tag{16}$$

**ix. HappyCat** [21] (interval: [-100, 100])

$$f_9(x)=\left|\sum_{i=1}^{d}x_i^2-d\right|^{1/4}+\left(0.45\sum_{i=1}^{d}x_i^2+\sum_{i=1}^{d}x_i\right)\Big/d+0.5 \tag{17}$$

**x. Dixon-Price** [22] (interval: [-10, 10])

$$f_{10}(x)=\sum_{i=1}^{d}(x_i-1)^2+\sum_{i=1}^{d-1}i\left(2x_{i+1}^2-x_i\right)^2 \tag{18}$$

### 3.2. Performance criteria

In order to evaluate the performance of metamodeling techniques, three criteria in Table 1 are calculated.

**Table 1** Criteria for performance evaluation.
where STD stands for standard deviation, MSE (Mean Square Error) represents the departure of the metamodel from the real simulation model, and the variance captures the irregular of the problem.

| Criteria | Expression |
|---|---|
| **Relative Average Absolute Error (RAAE)** | $\sum_{i=1}^{n}\lvert y_i-\hat{y}_i\rvert/(n\cdot\mathrm{STD})$ |
| **Relative Maximum Absolute Error (RMAE)** | $\max(\lvert y_i-\hat{y}_i\rvert)/\mathrm{STD}$ |
| ***R* square (*R*²)** | $1-\frac{\sum_{i=1}^{n}(y_i-\hat{y}_i)^2}{\sum_{i=1}^{n}(y_i-\bar{y}_i)^2}=1-\frac{\mathrm{MSE}}{\mathrm{variance}}$ |

### 3.3. Results and discussions

3.3.1. Comparisons between different metamodel techniques.

The KPLS in Ref. [4] is competent in both simulating the implement of dynamic simulation accurately and estimating the error of the predictor efficiently. The RBF-HDMR in Ref. [1] also works well for high-dimensional problems. Therefore, the KPLS and RBF-HDMR have been employed and compared with the CNN-based metamodel.

In Ref. [4], a strongly nonlinear problem, calculated by Eq. (19), had been tested with $d$=30, 50, 100, 150, 200, 250 and 300. Since this study deals with over hundred-dimensional problems, only the results of $d$=100, 150, 200, 250 and 300 are reported in Fig. 2.

$$f(x)=\sum_{i=1}^{d}\left[\left(x_i^2\right)^{x_{i+1}^2+1}+\left(x_{i+1}^2\right)^{x_i^2+1}\right],\ 0\le x_i\le 1 \qquad (19)$$

According to Fig. 2, $R^2$ and RAAE of the CNN-based metamodel are better than the ones of RBF-HDMR while the RMAE is worse. A small RMAE is preferred. While a large RMAE indicates a large error even though the overall accuracy indicated by $R^2$ and RAAE can be very good. However, since RMAE cannot show the overall performance in the design space, it is not as important as $R^2$ and RAAE [23]. The larger the $R^2$ is, the more accurate the metamodel is. Therefore, with the dimension is up to more than 100, the CNN-based metamodel might have a better performance.

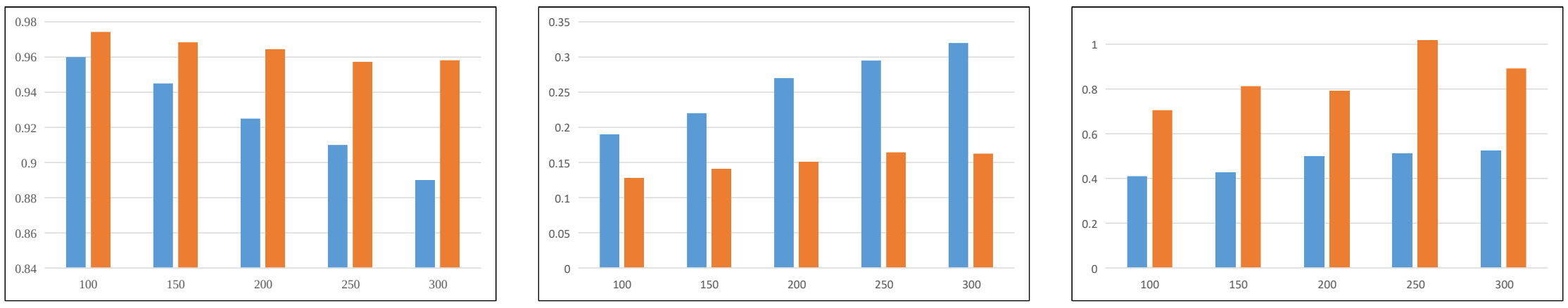

(a) $R^2$ (b) RAAE (c) RMAE

**Fig. 2.** Performance comparisons between the CNN-based metamodel and RBF-HDMR. where CNN-based metamodel is marked in orange and RBF-HDMR is marked in blue.

Bouhlel employed KPLS for high-dimensional problems. In Ref. [1], the dimension of Griewank function was set to 5, 7, 10, 20, 60 and interval was [-600, 600]. The following relative error calculated by Eq. (20) evaluated the performance of the KPLS model.

$$\text{Error} = \frac{\left\| \hat{\mathbf{Y}} - \mathbf{Y} \right\|_2}{\left\| \mathbf{Y} \right\|_2} \times 100\% \tag{20}$$

where $\left\| \bullet \right\|_2$ represents the usual $L_2$ norm.

As shown in Ref. [1], the Error of the KPLS model was satisfied within 10 dimensions (less than 1). However, when the dimension was increased to dozens, the Error of the KPLS had a significant increase (more than 5). For the CNN-based metamodel, as shown in Table 2, the Error doesn't change a lot even if the dimension is increased to 784. It suggests that the CNN-based metamodel might outperform others for more than 100 dimensional problems.

**Table 2** Error for the Griewank function in the interval [-600, 600] by the CNN-based metamodel.

| **Dimension** | **100d** | **144d** | **256d** | **324d** | **784d** |
|---|---|---|---|---|---|
| **Error** (%) | 0.921 | 0.972 | 1.241 | 1.211 | 1.225 |

3.3.2. Performances of the CNN for analytical examples.

As shown in Fig. 3, some other 324-dimensional* mathematical examples from Section 3.1 are further tested. It can be seen that most $R^2$ are more than 0.8, while the worst one is also more than 0.6. In our opinion, the $R^2$ might be satisfied considering the dimension is hundreds. The maximum (worst) RAAE is 0.5160 and the maximum (worst) RMAE is 2.8487. It can improve that the CNN-based metamodel might model well for more than 100 dimensional and strongly nonlinear problems.

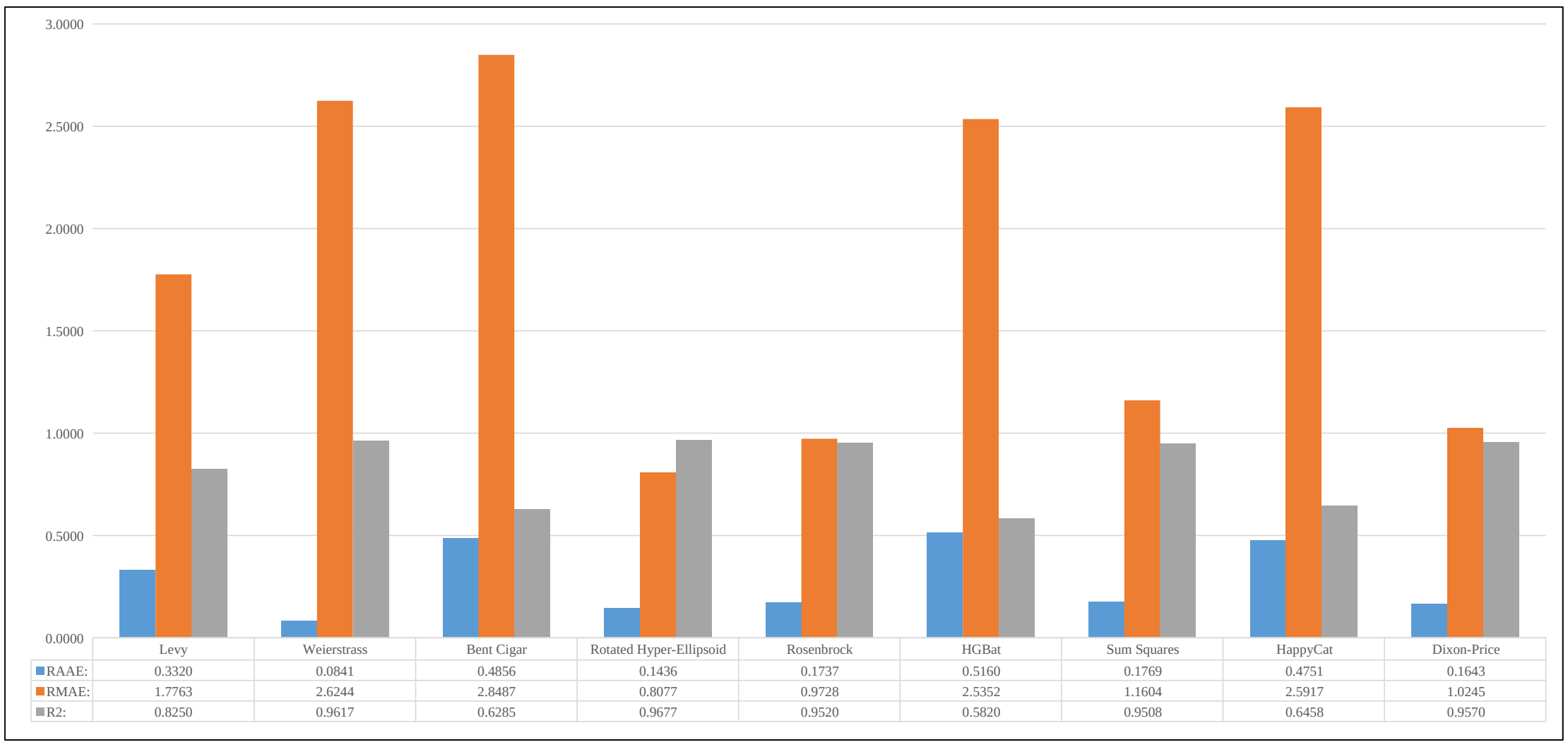

**Fig. 3.** The computational results of the test functions by the CNN-based metamodel.

3.3.3. Performances of the CNN for asymmetric example.

To make sure that all analytical examples are not highly symmetric about the origin, the expanded Griewank's plus Rosenbrock's function, expressed by Eq. (21), is modeled. The dimension is 324, and the interval is [-1, 1].

$$f_{11}(x)=f_1\left(f_6(x_1,x_2)\right)+f_1\left(f_6(x_2,x_3)\right)+\ldots+f_1\left(f_6(x_d,x_1)\right) \tag{21}$$

After calculation, RAAE is 0.169, RMAE is 0.994, $R^2$ is 0.955, and Error is 1.98%. The results demonstrate that the CNN-based metamodel might be powerful for unsymmetrical functions.

## 4. Geometric Modeling by the CNN

In the traditional design-and-analysis procedure, the geometric design and

* 324 is just a selected dimension number which is far larger than existing levels, and it is not based on a practical problem.

physical analysis are commonly considered as completely different engineering fields, and the gap between Computer Aided Design (CAD) and Computer Aided Engineering (CAE) models always exists. In order to integrate CAD and CAE models seamlessly, Hughes [24] proposes a CAD/CAE integration method named IsoGeometric Analysis (IGA). In this section, a model evaluated by IGA is employed to be modeled based on the CNN.

### 4.1. The bases of isogeometric design

It is well known that Non-Uniform Rational B-spline (NURBS) is a basic geometric design of the IGA, and its basis function can be mapped by the B-spline function.

$$R_{i,j}^{p,q}(\xi,\eta)=\frac{N_i(\xi)M_j(\eta)\omega_{i,j}}{\sum_{\hat{i}=1}^{n}\sum_{\hat{j}=1}^{m}N_{\hat{i}}(\xi)M_{\hat{j}}(\eta)\omega_{\hat{i},\hat{j}}} \tag{22}$$

where $p$, $q$ are the polynomial orders in different dimensions, $N_i(\xi)$ and $M_j(\eta)$, whose numbers are $n$ and $m$ respectively, are the standard B-spline basis functions in two dimensions, and $\omega_i$ is the weight value of each control point.

Give a set of control points $\mathbf{P}_i$ and a combination of bivariate NURBS basis functions, NURBS surface can be defined as

$$\mathbf{S}(\xi,\eta)=\sum_{i=1}^{n}\sum_{j=1}^{m}\mathbf{P}_{i,j}R_{i,j}^{p,q}(\xi,\eta) \tag{23}$$

When the parametric model is constructed, the displacement of the IGA can be obtained as

$$\mathbf{u}(\xi,\eta)=\sum_{i=1}^{n}\sum_{j=1}^{m}R_{i,j}^{p,q}(\xi,\eta)\mathbf{d} \tag{24}$$

$$\delta\mathbf{u}=\sum_{i=1}^{n}\sum_{j=1}^{m}R_{i,j}^{p,q}(\xi,\eta)\delta\mathbf{d} \tag{25}$$

where $\mathbf{d}$ and $\delta\mathbf{d}$ are the displacement vector at control points and displacement differential vector, respectively.

The displacement $\mathbf{d}$ in Eq. (24) can be calculated by

$$\mathbf{K}_1\mathbf{d}_1 = \mathbf{F} \tag{26}$$

where $\mathbf{K}_1$ is the stiffness matrix, and $\mathbf{F}$ is assumed constant load.

If Eq. (26) is discretized after a modification of the given initial structure, the $\mathbf{d}_1$ can be expressed as

$$\mathbf{d}_1 = \mathbf{d}_0 + \Delta\mathbf{d} \tag{27}$$

where $\Delta\mathbf{d}$ is the displacement increment, $\mathbf{d}_0$ is the initial displacement.

Substitute Eq. (27) into Eq. (26) as

$$\mathbf{K}_1\Delta\mathbf{d} = \boldsymbol{\delta};\ \boldsymbol{\delta} = \mathbf{F} - \mathbf{K}_1\mathbf{d}_0 \tag{28}$$

where $\boldsymbol{\delta}$ is the residual of initial solution.

The displacement increment is approximately defined by a linear combination of basis vector of influenced Degree of Freedoms (DOFs) as

$$\Delta\mathbf{d} = y_1\mathbf{v}_1 + y_2\mathbf{v}_2 + \cdots + y_{n_d}\mathbf{v}_{n_d} = \mathbf{r}_B\mathbf{y} \tag{29}$$

Substitute Eq. (29) into Eq. (28) with multiplying $\mathbf{r}_B{}^T$ in both sides. Equation (24) is changed to

$$\mathbf{K}_R\mathbf{y} = \mathbf{F}_R \tag{30}$$

s.t.

$$\mathbf{K}_R = \mathbf{r}_B{}^T\mathbf{K}_1\mathbf{r}_B;\ \mathbf{F}_R = \mathbf{r}_B{}^T\boldsymbol{\delta} \tag{31}$$

After solving $\mathbf{y}$ in Eq. (30), substitute it into Eq. (29) to obtain the approximation $\triangle\mathbf{d}$. Ultimately, $\mathbf{d}_1$ can be obtained by Eq. (27).

For linear elasticity problems, the stress vector can be defined as

$$\boldsymbol{\sigma}_e = \mathbf{D}\delta\mathbf{d}_e = \mathbf{D}\mathbf{B}\mathbf{d}_e \tag{32}$$

where $\mathbf{D}$ is the elasticity matrix, $\mathbf{B}$ is the strain-displacement matrix, and $\mathbf{d}_e$ is the displacement vector in the corresponding element.

## 4.2. IGA model

In this study, Rhinoceros software is used to provide the input information of a tubular structure which is shown in Fig. 4. As shown in Figs. 5(b) - (c), the DOFs in the bottom of the structure are fixed. In addition, as shown in Fig. 5(a), several concentrated forces $3\times10^5 N$ in $y$ direction are enforced to the control points of the tubular top.

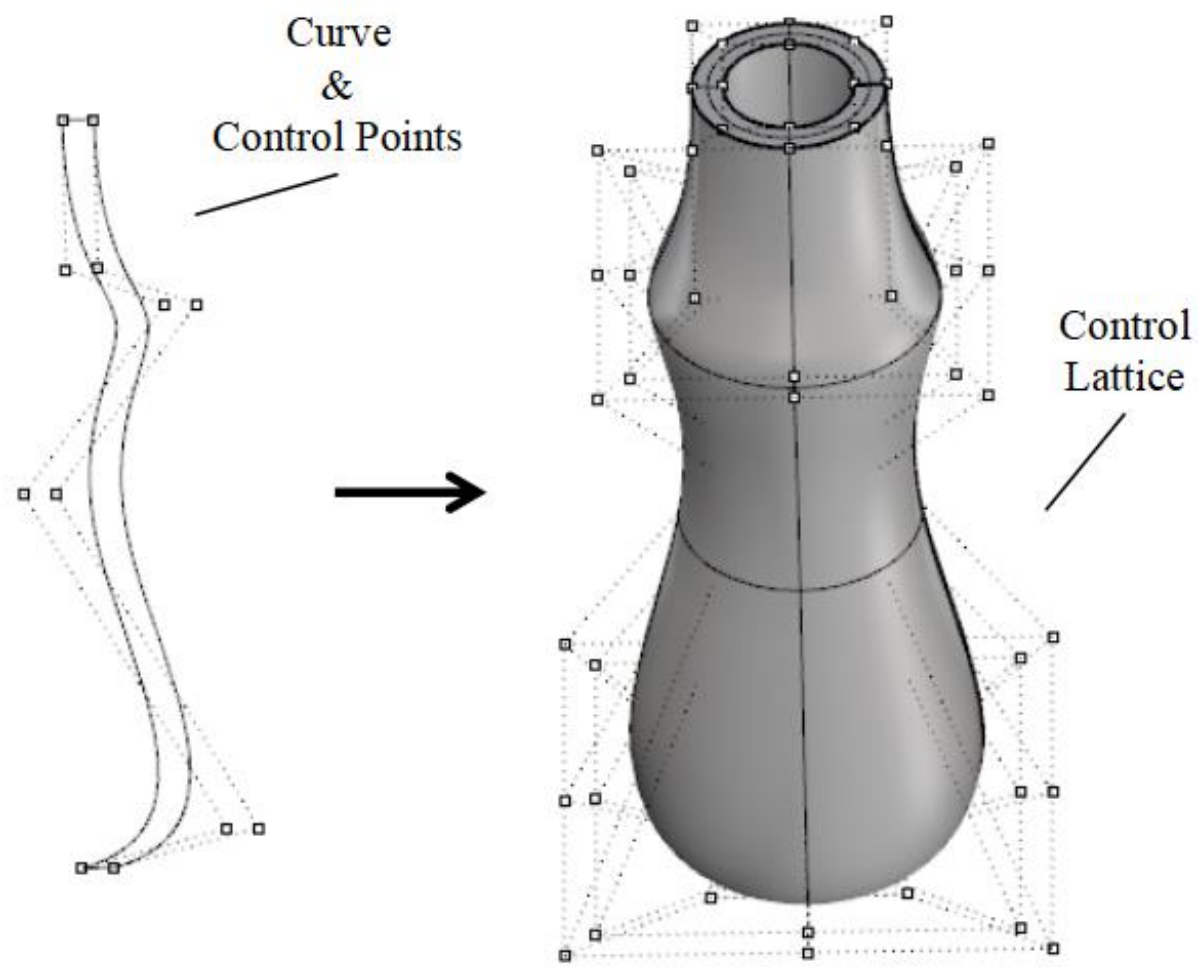

**Fig. 4.** The control lattice for the tubular structure based on IGA.

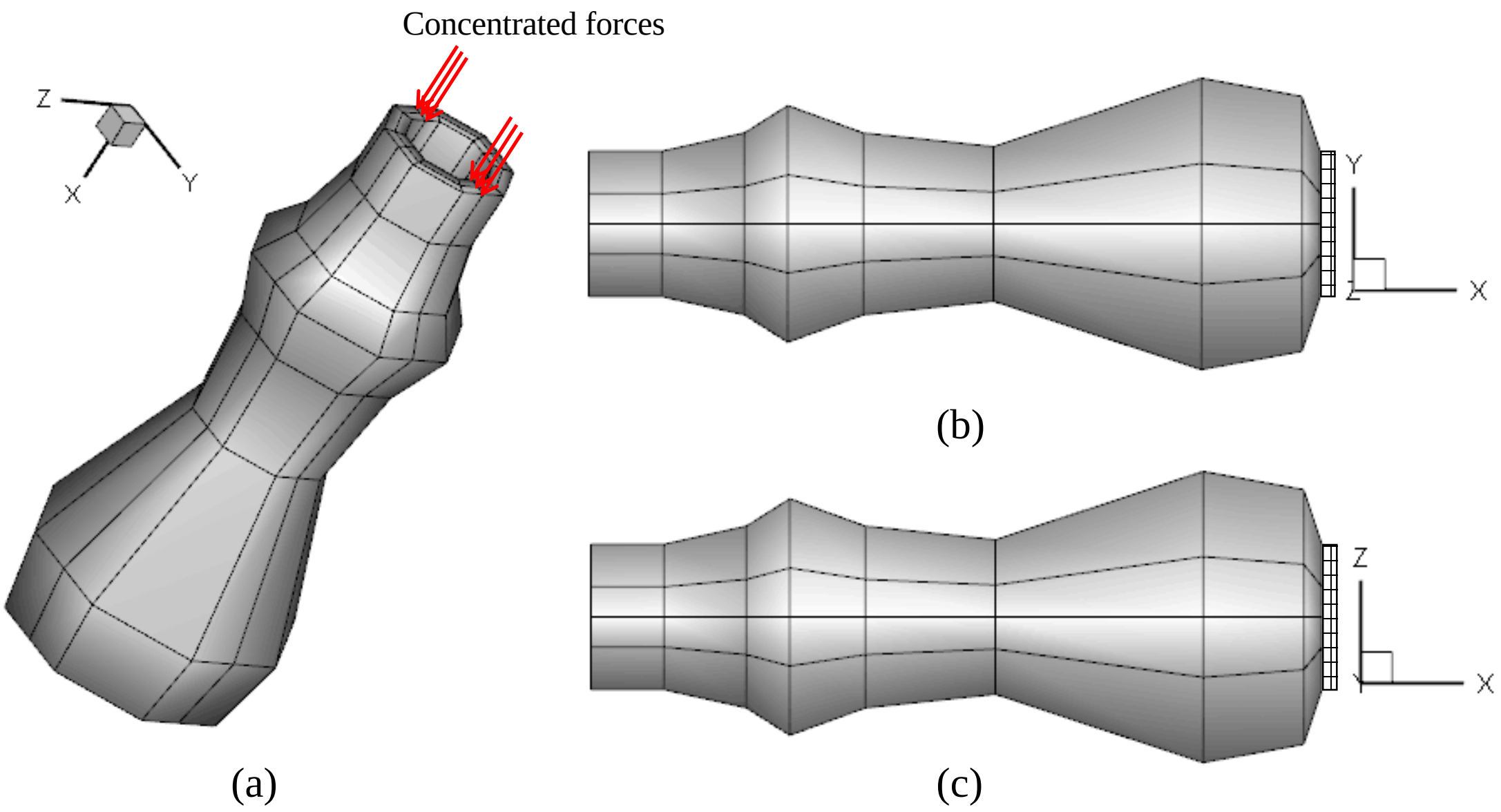

**Fig. 5.** The constraints and load of the tubular structure.

### 4.3. CNN-based metamodel

The control points are used as the design variables of each sample for the CNN and the maximum stress is the learning label. For each sample, design variables include 324 control points, and each control point includes three coordinates ($x_i$, $y_i$, $z_i$) and a weight factor $w_i$. However, the tubular structure particularly changes only in the direction of the $y$-axis, and each $w_i$ is fixed. Therefore, each sample contains 324 $y_i$s.

Each $y_i$ changes randomly in the interval [$0.8y_{i\text{-}init}$, $1.2y_{i\text{-}init}$], and 10,000 training and 10,000 testing samples are obtained. The training results are shown in Table 3. It can be found that the RAAE and the RMAE are small enough. As the $R^2$ is 0.997, it indicates that the trained CNN-based metamodel might perform well.

**Table 3** Criteria for the CNN-based metamodel for the IGA application.

| **RAAE** | **RMAE** | $\boldsymbol{R^2}$ |
|---|---|---|
| 0.033447 | 0.52820 | 0.99776 |

Then the constructed CNN-based metamodel is optimized by Particle Swarm Optimization (PSO). The objective function to be minimized is the maximum stress in the tubular structure, and the $y_i$s of control points are chosen as the design variables. The optimization problem is stated as

$$\min\ f = CNN\left(\mathbf{B}\left(y_1, y_2, \ldots, y_{324}\right)\right) \tag{33}$$

s.t.

$$y_i \in \left[0.8y_{i-init},\ 1.2y_{i-init}\right],\ i \in 1, \ldots, 324 \tag{34}$$

where $CNN(\mathbf{x})$ is the maximum stress in the tubular structure, $\mathbf{B}$ is the control points, and $y_{i\text{-}init}$ is the initial value of each $y_i$.

The comparison of optimization processes between the CNN-based metamodel and the IGA is shown in Fig. 6. The optimal stress distribution by the CNN-based metamodel is shown in Fig. 7. The optimum by the CNN-based metamodel is 1,340,470$N$ and the optimum by the IGA is 1,354,991$N$. It can be seen that the optimum by the CNN-based metamodel is 1.072% less than the optimum by the IGA. The optimization represents that the CNN-based metamodel might be a potential feasible way for real engineering problems.

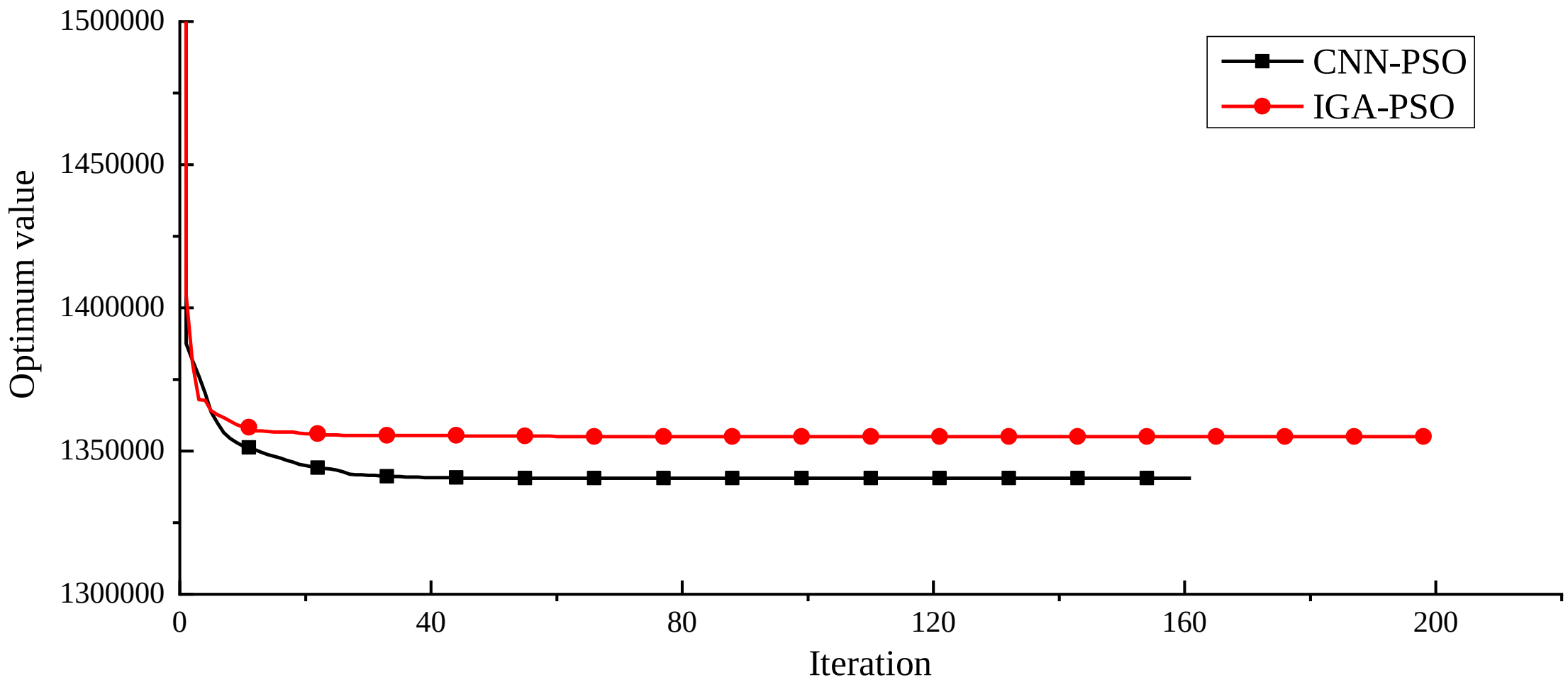

**Fig. 6.** The comparison between optimizations of the CNN-based metamodel and IGA for the IGA optimization.
where the population of the PSO is 50.

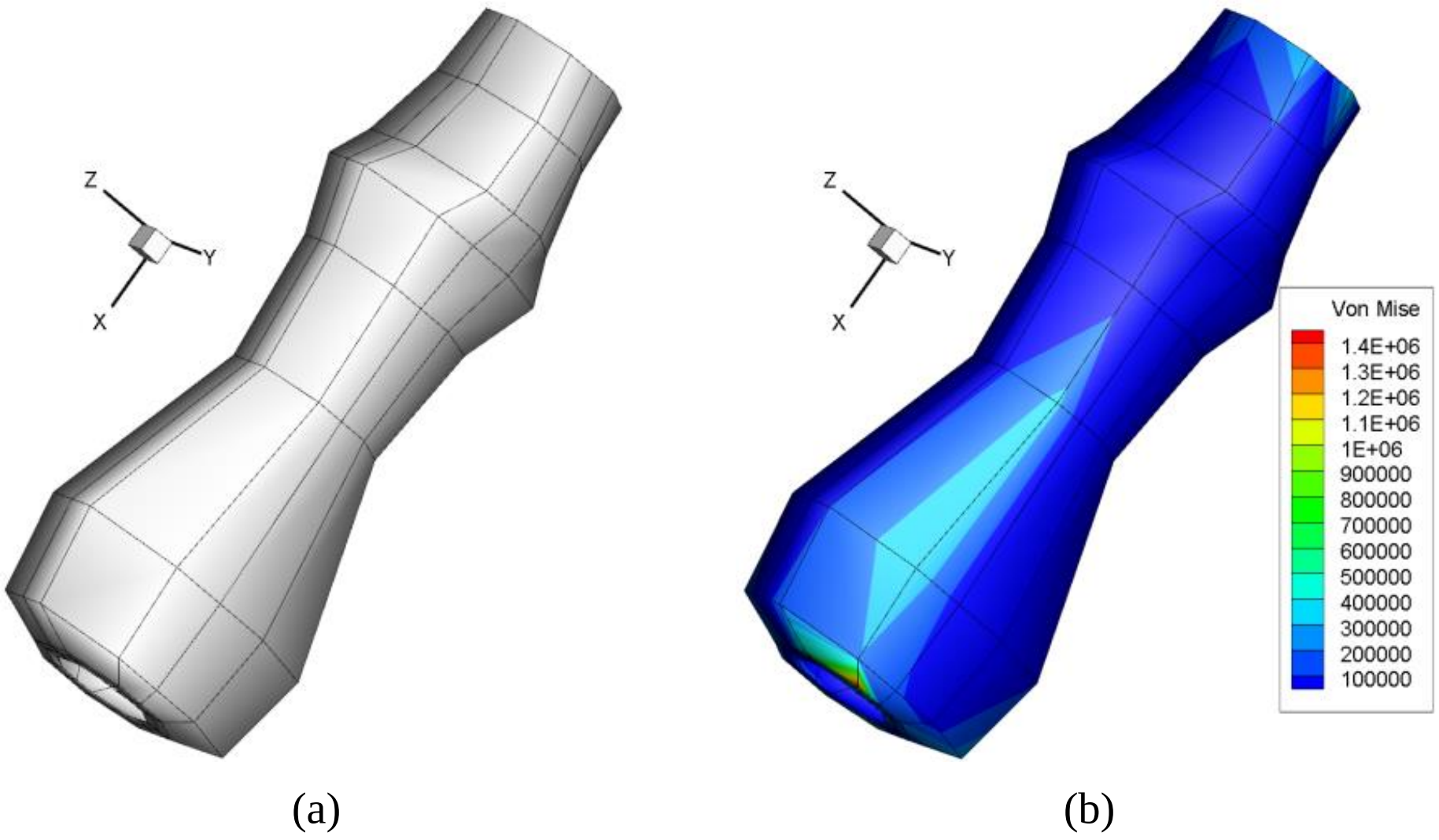

**Fig. 7.** The optimal shape and stress distribution of the optimal structure by the CNN-based metamodel.

## Conclusions

This letter attempts the CNN-based metamodel technique to model for more than one hundred-dimensional and strong-nonlinear problems. In order to evaluate the CNN-based metamodel, the RBF-HDMR and KPLS mentioned in Refs. [4] and [1] respectively have been employed and compared. The results demonstrate that the CNN might perform well for high-dimensional problems. Subsequently, to further analyze the CNN-based metamodel, several high-dimensional and strong-nonlinear analytical functions are modeled by the CNN and the modeling results are satisfied.

Finally, the CNN is applied to an IGA-based optimization application. The results indicate that the CNN-based metamodel might be a potential feasible way for real engineering problems.

## Acknowledgment

This work has been supported by Project of the National Key R&D Program of China 2017YFB0203701 and Key Program of National Natural Science Foundation of China under the Grant Numbers 11572120.